\documentclass[border=2pt]{article}

\usepackage[utf8]{inputenc} 
\usepackage[T1]{fontenc}    
\usepackage{hyperref}       
\usepackage{url}            
\usepackage{booktabs}       
\usepackage{amsfonts}       
\usepackage{nicefrac}       
\usepackage{microtype}      

\usepackage{xcolor}
\usepackage{graphicx}
\usepackage{grffile}
\usepackage{mathtools}
\usepackage{ragged2e}
\usepackage{amssymb}
\usepackage{caption}
\usepackage{algpseudocode}
\usepackage{makecell}
\usepackage{comment}
\usepackage{tikz}
\usepackage{lipsum,adjustbox}
\usepackage{algorithm,algpseudocode}
\usepackage{amsthm}
\usepackage{lscape}
\usepackage{subfigure}
\usepackage{pdflscape}
\usepackage{amssymb}
\usepackage{pifont}
\usepackage{amsfonts}
\usepackage{tkz-euclide}
\usepackage{pythonhighlight}
\usepackage{multirow}
\usepackage{multicol}
\usepackage[english]{babel}
\usepackage[leftcaption]{sidecap}
\usepackage{amsmath}

\usepackage{paralist,pst-func, pst-plot, pst-math, pstricks-add,pgfplots}
\usetikzlibrary{patterns,matrix,arrows}

\definecolor{applegreen}{rgb}{0.55, 0.71, 0.0}
\definecolor{oceanboatblue}{rgb}{0.0, 0.47, 0.75}
\definecolor{paleviolet-red}{rgb}{0.86, 0.44, 0.58}
\definecolor{goldenyellow}{rgb}{1.0, 0.87, 0.0}
\definecolor{fluorescentorange}{rgb}{1.0, 0.75, 0.0}

\newcommand \FigPath[1]{#1}
\newcommand \FigRef[1]{Figure \ref{#1}}
\newcommand \TableRef[1]{Table \ref{#1}}
\newcommand \AlgoRef[1]{Alg. \ref{#1}}

\usetikzlibrary{positioning,shapes.symbols, arrows, calc}

\tikzset{dexteritas/.cd,
	shifted path label/.style={pos=0.5,draw=none,rectangle,auto,sloped}}
\tikzset{shifted path/.style args={from #1 to #2 by #3}{insert path={
			let \p1=($(#1.east)-(#1.center)$),
			\p2=($(#2.east)-(#2.center)$),\p3=($(#1.center)-(#2.center)$),
			\n1={veclen(\x1,\y1)},\n2={veclen(\x2,\y2)},\n3={atan2(\y3,\x3)} in
			(#1.{\n3+180+asin(#3/\n1)}) to 
			(#2.{\n3-asin(#3/\n2)})
}}}

\tikzset{labeled shifted path/.style args={from #1 to #2 by #3 label #4}{insert path={
			let \p1=($(#1.east)-(#1.center)$),
			\p2=($(#2.east)-(#2.center)$),\p3=($(#1.center)-(#2.center)$),
			\n1={veclen(\x1,\y1)},\n2={veclen(\x2,\y2)},\n3={atan2(\y3,\x3)} in
			(#1.{\n3+180+asin(#3/\n1)}) to node[dexteritas/shifted path label]{#4}
			(#2.{\n3-asin(#3/\n2)})
}}}

\newcommand \EquationRef[1]{Eq. \ref{#1}}
\newcommand \SectionRef[1]{Sec. \ref{#1}}

\newcounter{MyAlgorithm}
\newcounter{PlaceHolder}

\graphicspath{{./picture/}} 

\title{
Joint Manifold Learning and Density Estimation Using Normalizing Flows
}

\author{%
  Seyedeh Fatemeh Razavi, Mohammad Mahdi Mehmanchi,
  \\
  Reshad Hosseini, Mostafa Tavassolipour\\
  \{razavi\_f, mahdi.mehmanchi, reshad.hosseini, tavassolipour\}@ut.ac.ir
  \\
  University of Tehran
  \\
  Tehran, Iran
}
\begin{document}

\maketitle


\begin{abstract}
    Based on the manifold hypothesis, real-world data often lie on a low-dimensional manifold, while normalizing flows as a likelihood-based generative model are incapable of finding this manifold due to their structural constraints.
    So, one interesting question arises:
    \textit{"Can we find sub-manifold(s) of data in normalizing flows and estimate the density of the data on the sub-manifold(s)?"}.
    In this paper, we introduce two approaches, namely per-pixel penalized log-likelihood and hierarchical training, to answer the mentioned question.
    We propose a single-step method for joint manifold learning and density estimation by disentangling the transformed space obtained by normalizing flows to manifold and off-manifold parts.
    This is done by a per-pixel penalized likelihood function for learning a sub-manifold of the data.
    Normalizing flows assume the transformed data is Gaussianizationed, but this imposed assumption is not necessarily true, especially in high dimensions.
    To tackle this problem, a hierarchical training approach is employed to improve 
    the density estimation on the sub-manifold.
    The results validate the superiority of the proposed methods in simultaneous manifold learning and density estimation using normalizing flows in terms of generated image quality and likelihood.
\end{abstract}

\section{Introduction}
%
There are a variety of well-known deep likelihood-based generative methods, like Variational Autoencoders (VAEs) \cite{kingma2013auto}, Normalizing Flows (NFs) \cite{rezende2015variational}, Auto-Regressive (AR) models  \cite{murphy2022probabilistic}, Energy Base (EB) models \cite{murphy2022probabilistic}, and Diffusion Models (DMs) \cite{pmlr-v37-sohl-dickstein15}.
But among the mentioned models,
only ARs and NFs can exactly compute the likelihood. VAEs and DMs find a lower bound of the likelihood, and EB models approximate it.
Sampling in common AR models is computationally expensive, due to sequential nature of these models.
Sampling in NFs is not sequential, but they have a structural limitation that limits their applicability and generating power.


The important structural limitation of common NFs is that they cannot learn the embedded sub-manifolds of the data. Real data are usually embedded in a low-dimension sub-manifold, and powerful generating models make use of this assumption. NFs use bijective transformation, and therefore they preserve dimensionality between the input and transformed spaces. One can think that this can be easily solved by using non bijective transformation that can transform the data space to a lower dimensional space, and maximizing the likelihood of the data on the low-dimensional sub-manifold obtained by such a transformation. But unfortunately, this optimization problem can not be solved exactly. There have been several attempts to solve this problem, but they have shortcomings.

Recently, several researchers proposed solutions to the problem of maximizing the likelihood of the data on a sub-manifold using NFs.
Some use an injecting transformation in NFs \cite{brehmer2020flows, caterini2021rectangular, cunningham2020normalizing}.
By using injective transformations, non-square Jacobian matrices appear in the log-likelihood making the optimization very computationally expensive.
The authors in \cite{brehmer2020flows} used two-step training procedure, first the manifold is learned using a NF, then the density is estimated using another NF. This procedure simplifies the training, but it can lead to poor density estimation.
Overcoming the two-step training with a linear algebra trick in an optimization method was proposed in \cite{caterini2021rectangular}.
However, it is computationally expensive in high dimensions.
In \cite{horvat2021denoising}, the authors introduced a method that learns a low-dimensional representations from NFs in a single-step training procedure without using injective transformations.
The transformed space of the NF is split into two parts, manifold and noise, modeled by another NF and a low-variance Gaussian distribution, respectively. They also add this low-variance noise to the input. We use the structure similar to \cite{horvat2021denoising} but with a different loss function and training procedure, which leads to significantly better results. Before stating the key components of our procedure leading to its success, we discuss the objective function of likelihood-based models which is the main reason of their inferior generating power compared with non-likelihood-based models.

\begin{figure}[H]
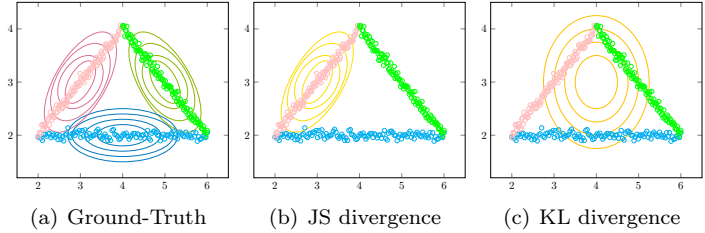

    \subfigure[Ground-Truth]
    {\begin{adjustbox}{width=0.25 \textwidth}
    		\input{GT1.tikz}
    \end{adjustbox}}
    \subfigure[JS divergence]
    {\begin{adjustbox}{width=0.25 \textwidth}
    		\input{JS1.tikz}
    \end{adjustbox}}
        \subfigure[KL divergence]
    {\begin{adjustbox}{width=0.25 \textwidth}
    		\input{KL1.tikz}
    \end{adjustbox}}
    \centering
	\caption{Illustration of a Gaussian density fitted to  data by different distance measures.
The true underlying density can be represented relatively well by multiple 1D manifolds.
}
\label{fig:distance}
\end{figure}

The aforementioned likelihood-based generative methods such as NFs optimize the Kullback-Leibler (KL) divergence between the model and the true underlying distribution.
They assign equal weights to samples. Consequently, the model spreads the density to learn nearly all modes.
Therefore, minimizing KL-divergence (or equally maximizing likelihood) over the entire high-dimensional space is challenging \cite{theis2015note}.
As a result, generated samples in these methods have high diversity and are prone to low quality.
In contrast, non-likelihood-based generative models, like Generative Adversarial Network (GAN) \cite{goodfellow2014generative},
minimize the Jensen-Shannon (JS) divergence between the model and the true underlying distribution.
They can learn fewer modes of the data density well and generate samples with higher quality. 
The graphical intuition of the above discussion is presented in \FigRef{fig:distance}.


To address the main question of the current research, that is joint learning the density and a submanifold of data, we propose two approaches: finding a good sub-manifold in NFs by penalizing the loss function and better estimating the likelihood of the sub-manifold by a hierarchical method.

\begin{figure}[H]
    {\includegraphics[width=0.75\linewidth]{\FigPath{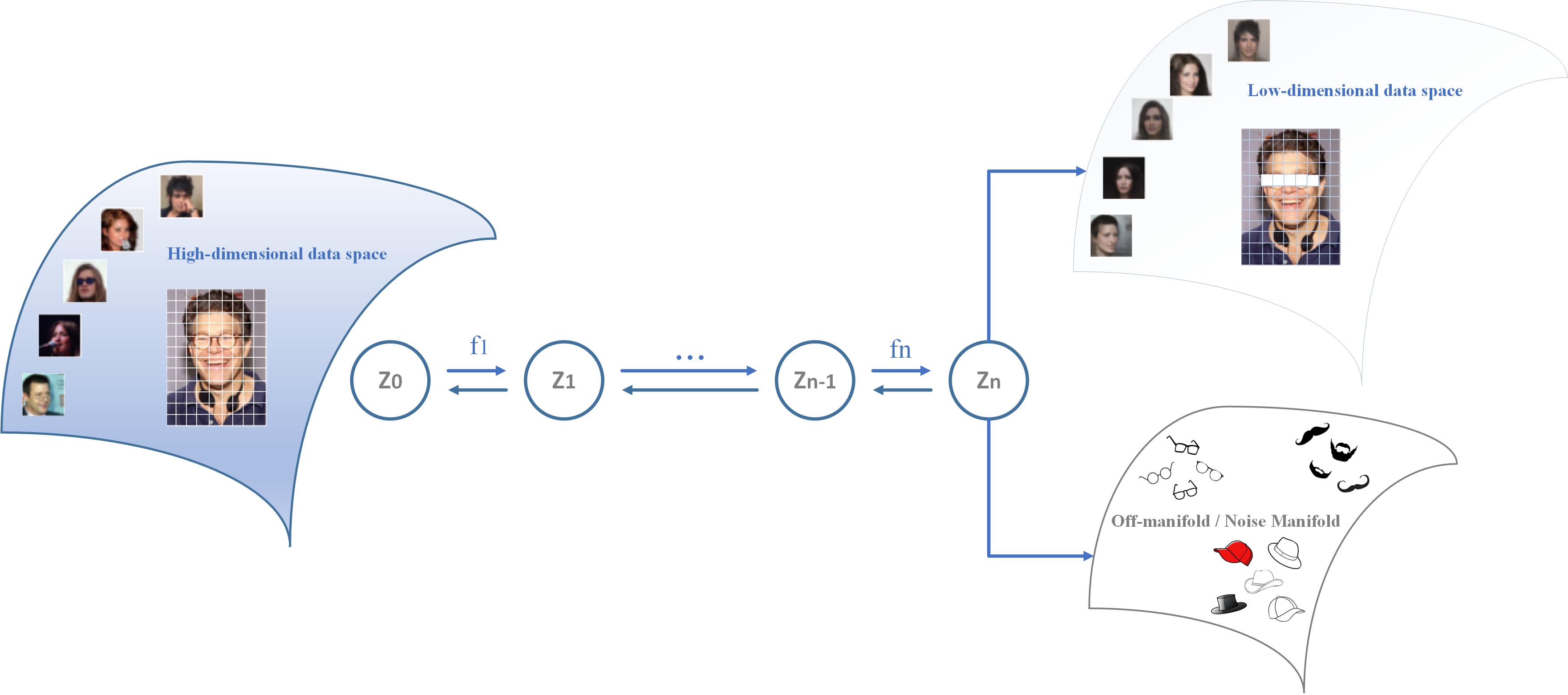}}}
    \centering
	\caption{
Pixel rejection method.
The last transformed space by an NF is disentangled into a manifold part (pixels belong to facial structure) and an off-manifold part (pixels belong to objects that cause facial variations).
	}
	\label{fig:PixelRejection}
\end{figure}

The first method overcomes the aforementioned inherent shortcoming of NFs as a likelihood-based generative model in order to estimate a better sub-manifold, particularly in high dimensions.
Estimating such a sub-manifold is done by penalizing the usual loss function of NFs with a per-pixel Huber function.
In this approach that we call pixel rejection, some pixels are considered to lie on the main sub-manifold and the rest are considered to be off-manifold.
Using per-pixel Huber function, the manifold pixels are penalized more than off-manifold pixels.
Therefore, the inverse mapping is trained to reconstruct the original data using the main sub-manifold of data. 
For a better understanding, the main idea of the pixel rejection method for an image data is shown in \FigRef{fig:PixelRejection}.
Briefly, the transformed data by an NF can be disentangled into the manifold and off-manifold spaces.
The main structure of the face is considered the main manifold, and other external factors (like skin color, beard, glass, hats, background, etc.) can be considered as off-manifold. 

The second method to answer the main question of this study is providing a hierarchical training approach for a more accurate likelihood estimation of the sub-manifold. NFs actually learn a bijective transformation for Gaussianizing the data, meaning that the transformed data have Gaussian density. When the dimensionality of the data is high, the transformed space would be far from Gaussian, because of the KL-divergence problem we discussed beforehand. We tackle this problem with a hierarchical learning procedure on a hierarchical NF structure. 
In the first level of the hierarchy, we jointly learn the sub-manifold of the data and the density of the data. But since the density is fitted on the whole space, the dimensionality is high and we do not have a good fit. In the next step of the hierarchy, the learned manifold from the first step is considered as the input for the NF of the next step. Therefore, the density is fitted on a much lower dimension (i.e. the dimensionality of the resulting sub-manifold of the first stage) and the transformed density of the second stage would be much closer to Gaussian. This procedure can be repeated in several stages, and in each stage we can reduce the dimensionality of the sub-manifold. Therefore, this hierarchical training strategy pays more attention to fitting a Gaussian density on the transformed low-dimensional sub-manifold than on the transformed whole space, which we know it can not be well fitted. Based on the intuition of this structure and training procedure, it is expected to have better generating power.

The paper is arranged as follows.
\SectionRef{sec:related_works} reviews the recently proposed methods related to 
manifold learning
in NFs. 
The
prerequisites are introduced in \SectionRef{sec:preliminaries}.
\SectionRef{sec:proposed_method} introduces the proposed method.
Experimental results are presented in \SectionRef{sec:results}.
In the end, the conclusion is discussed in \SectionRef{sec:conclusion}.
\section{Related Works}
\label{sec:related_works}
\textbf{Flows on prescribed manifolds}
Some
studies
considered learning flows on manifolds that are already known \cite{gemici2016normalizing, mathieu2020riemannian}.
Mainly, flows on tori and sphere spaces are learned in \cite{rezende2020normalizing}. When the manifold is known, computation of the density’s log Jacobian determinant may be easy, but this method can only be used when the data manifold is known beforehand.

\textbf{Flows on learnable manifolds}
Recently, solutions to the problem of manifold learning and density estimation with NFs have been proposed \cite{kim2020softflow, kothari2021trumpets, cunningham2020normalizing,brehmer2020flows,caterini2021rectangular,horvat2021denoising,kalatzis2021multi,ross2021conformal}.
In the following, we review the most related works to our research and highlight their key properties.
A proposed scientific path is about separating manifold learning and density estimation.
The pioneer of this study in NF is a method named M-flow \cite{brehmer2020flows}.
In M-flow, an injective transformation is used to transform the transformed space of an NF into a lower dimensional space (a sub-manifold). After that, another NF is used to estimate the density on the sub-manifold. 
Manifold learning and density estimation are done separately to avoid the computation of Jacobian term induced by injective transformation (two-phase training).
An extension of this study called multi-chart flow employs multiple mapping instead of one \cite{kalatzis2021multi}. This paper aims to find a multiple-chart manifold, and also suffers from two-phase training.
Another followed research of M-flow, named Rectangular flows \cite{caterini2021rectangular},
overcomes the calculation of injective transformation by relying on automatic differentiation and linear algebra tricks.
A recent study, named Denoising Normalizing Flow (DNF) \cite{horvat2021denoising} overcomes the limitation of using injective transformation and separating model training by splitting the transformed space of an NF into two parts, noise-insensitive and noise-sensitive.
These parts are modeled by another NF and a low-variance Gaussian distribution, respectively. The noise is also added to the input data. 
Using this structure, two-phase training is no longer needed and the NFs are trained simultaneously.
\section{Preliminaries}
\label{sec:preliminaries}
This section introduces our notations and related preliminary works to make it easier for the reader to follow the subject.
The rest of this section is arranged as follows:
At first, the standard normalizing flow is discussed in \SectionRef{sec:nf}.
After that, a robust reconstruction loss function is presented in \SectionRef{sec:reconstruction}.

\subsection{Normalizing Flow}
\label{sec:nf}
Normalizing flow is a parametric diffeomorphism transformation, $f_{\phi}:\mathbb{R}^D\rightarrow \mathbb{R}^D$.
So, it is a two-side differentiable bijective transformation.
By choosing a random variable from a pre-defined distribution in $\mathbb{R}^D$ and transferring it by $f_{\phi}^{-1}$, the distribution of transformed data can be found by a change-of-variable formula like 
\begin{equation}
    p_X(X) = p_Z(f_{\phi}(X)) \vert \det G(f_{\phi}(X)) \vert ^ {-\frac{1}{2}},
    \label{eq:nf}
\end{equation}
where $G \in \mathbb{R}^{D \times D}$ is defined by $G = J_{f_{\phi}^{-1}}^T J_{f_{\phi}^{-1}}$.
It is clear that the practical computability constraint for this term is the implicit constraint imposed on flow models.
All in all, the model parameters are estimated by the maximum log-likelihood criterion, like $\phi^{*} = \arg \underset{\phi}{\max} \log p_X(x)$
where $x=\{x_n\}_{n=1}^{n=N}$ is the available data from the distribution $p_X(x)$.

\subsection{Reconstruction loss functions}
\label{sec:reconstruction}
During the training of normalizing flow on a manifold, all off-manifold samples have a likelihood of 0.
Penalizing the off-manifold part through a quadratic reconstruction function (Mean Square Error (MSE)) is the common approach to tackle this problem in the literature
\cite{brehmer2020flows,
caterini2021rectangular,
horvat2021denoising,
ross2021conformal}.
A weakness of a quadratic function is its sensitivity to off-manifolds.
It switches between a linear and quadratic function based on comparing the error value with a threshold $\delta$.
Accordingly, this switching leads the
off-manifolds
are not penalized as much as 
on-manifold
data.
\EquationRef{eq:huber} defines the Huber function.
\begin{equation}
\label{eq:huber}
    H_\delta(x,y) = 
     \begin{cases}
      0.5(x-y)^2, &if |x-y| < \delta
      \\
      \delta(|x-y| - 0.5 \delta), & otherwise  \\ 
     \end{cases}
\end{equation}


\section{Proposed method}
\label{sec:proposed_method}
As mentioned earlier, the main goal of this paper is to answer to
\textit{"Is it possible to learn the likelihood of the embedded sub-manifold(s) in NFs?"} through two approaches.
At first, we propose a method in \SectionRef{sec:pixelrejection} to disentangle the transformed space of NFs to manifold and off-manifold.
After that, a hierarchical learning method is presented in \SectionRef{sec:hierarchical} that helps estimating the density on the sub-manifold, by making the transformed sub-manifold closer to the Gaussian density in several stages.

\subsection{Pixel rejection}
\label{sec:pixelrejection}
Likelihood-based generative models tend to learn all modes of data distribution because of minimizing the KL-divergence between the true underlying distribution and the model distribution.
Capturing all data variations, especially in high dimensional data, makes it difficult to estimate the density accurately by the model \cite{theis2015note}.
Relying on the manifold hypothesis, we pursue finding a dominant embedded low dimensional manifold of the data.
The data dominant sub-manifold is appraised as manifold space $\mathcal{M} \subseteq\mathbb{R}^{d}$, while its normal space is considered as off-manifold space $\mathcal{M} \subseteq\mathbb{R}^{D-d}$.
A common objective in the literature is to maximize the likelihood of data while penalizing the off-manifold components of the data.

Existing methods such as M-flow, DNF, and Rectangular flow penalize the whole data with a quadratic loss function.
An important characteristic of a quadratic function is its sensitivity to outliers, that can be interpreted as off-manifold parts of the data. 
Applying a linear or a quadratic loss to each components of data (which is pixel in images), depending on the distance to the manifold, is the main contribution of our proposed pixel rejection.
This switching is modeled with a Huber function.
In the pixel rejection method, an element-wise Huber function with threshold $\delta$, denoted $H_{\delta}$, is applied to the difference between the input data and the reconstructed one. The reconstructed data is a function of manifold rather than joint manifold and off-manifold parts.
%
More precisely, suppose that
$x = (x_1, x_2, ..., x_D)$
and
$\tilde{x} = f^{-1}(f_u(x)) = (\tilde{x}_1, \tilde{x}_2, ..., \tilde{x}_D)$
are the input data and its reconstruction (computed through the inverse of normalizing flow $f$), respectively.
Moreover, $f_u$ is the first $d^{th}$ components of the output of $f$ corresponding to the data manifold.
Therefore, the penalization term is computed by averaging the element-wise Huber function like $\mathcal{H}_{\delta}(x,\tilde{x}) = \frac{1}{D} \sum_{i=1}^{D} H_\delta(|x_i - \tilde{x}_i|)$.
Accordingly, a constrained optimization problem on Negative Log-Likelihood (NLL) appears like
\begin{equation}
    \underset{\phi}{\min} \;\textrm{NLL}(x) \;\textrm{s.t.} \; \mathcal{H}_{\delta}(x,\tilde{x}) \le \epsilon.
\end{equation}
Inspired by the Lagrange multipliers method, the proposed constrained optimization problem can be converted to its unconstrained equivalent one.
By penalizing the common objective function of NFs according to the mentioned method, we achieve a trade-off between the quality and the diversity of generated samples by varying $\delta$. With a higher $\delta$, generated sample diversity will increase, but the quality will decrease, and vice versa.
Therefore, the pixel rejection method is an algorithmic solution to overcome the problem of minimizing KL-divergence when data is embedded in a low-dimension manifold. 
It prevents generating poor samples by decreasing paying attention to all the space.

Details of the proposed method are as follows.
Let $f_\phi : \mathbb{R}^{D} \rightarrow \mathbb{R}^{D}$ be a standard NF. We denote the first $d$ components of the output of $f$ as $u$ and the remaining ones as $v$ where $u$ stands for on-manifold part of data and $v$ for off-manifold part. Formally,
$
z = f_\phi(x) = (z_1, z_2, ..., z_D),   u = (z_1, z_2, ..., z_d), v = (z_{d+1}, z_{d+2}, ..., z_D),
$
where $x$ is the input, and $z$ is the corresponding transformed data by the NF.
We choose $P_z(z) = N(0,I_D)$ and factorize $P_z(z)$ such that $P_z(z)=P_u(u)P_v(v)$.
Therefore, we have $P_u(u) = N(0,I_d)$ and $P_v(v) = N(0,I_{D-d})$.
Our goal is to re-arrange the transformed data of a normalizing flow in such a way that a part corresponds to a manifold and another one is off-manifold.
We achieve this goal by adding a 
penalization
term to our objective.
First we pad $u$ by a zero vector as 
$
\tilde{z} = (u, \vec{0}_{D-d}) 
$.
Our 
penalization
loss term is now defined using the Huber function as
\begin{equation}
r = \frac{1}{D}  \sum_{i=1}^{D} H_\delta(|x_i-f_\phi^{-1}(\tilde{z}_i)|),
\label{eq:reconstruction}
\end{equation}
where $H_\delta$ is a Huber function with threshold $\delta$.
All in all, our proposed loss function is defined as 
\begin{equation}
\ell = -\log P_u(u) - \log P_v(v) + \frac{1}{2}\log|\det G_{f_\phi}(x)| + \lambda r,
\label{eq:loss}
\end{equation}
where $\lambda$ is the
penalization
term hyper-parameter, and the first three terms
are the objective of a standard NF.
For generating sample, first we sample $u \sim P_u(u)$, then pad $u$ with $\vec{0}_{D-d}$ to obtain $\tilde{z}$, and eventually set $\tilde{x} = f_\phi^{-1}(\tilde{z})$.
The pseudo-code of the pixel rejection algorithm is given in \AlgoRef{algo:PixelRejection}.

In the end, it should be mentioned that we can use another NF $h_\theta$ to fit a more complex distribution on the sub-manifold instead of only the Gaussian distribution.
This idea was followed by DNF \cite{horvat2021denoising} and Mflow \cite{brehmer2020flows} to density estimation, too.
Briefly, in this case, $\log P_{u}(u)$ in \EquationRef{eq:loss} is replaced by $\log P_{u'}(u') + \frac{1}{2}\log|\det G_{f_\phi}(x)|$,
where $u'$ is the transformed space by $h_{\theta}$, and $P_{u'}(u') = N(0,I_d)$.
For generating sample it is enough to first we sample $u' \sim P_{u'}(u')$, then we compute $\tilde{x} = f_\phi^{-1}(h_\theta^{-1}(u'), \vec{0}_{D-d})$.
\begin{algorithm}[h]
	\caption{The pseudo-code of Pixel Rejection method.
	The used notation $\ell$ is computed as:
	\\
$\ell \gets
\frac{1}{N} \sum_{n=1}^{N} \bigg( -\log P_u(u_n) - \log P_v(v_n)
 + \frac{1}{2}|\det G_{f_\phi}(u_j,v_j)| 
 + \frac{\lambda}{D} \sum_{i=1}^{D} H_\delta(|x_{n,i} - \tilde{x}_{n,i}|)
 \bigg)$.
	}
	\label{algo:PixelRejection}
	\begin{multicols}{2}

\begin{algorithmic}[1]
		\Procedure{\underline{Training}}{}
		\\
            {\textbf{Require:}
            \small
			Manifold dimension d,
			Huber function threshold $\delta$, 
			Penalization term coefficient $\lambda$,
			Learning rate $\alpha$,
			data $X = \{x_n\}_{n=1}^{n=N} \sim p_X(x)$.
            }
			\While {has not converged}
			\For{$n \gets 1$ to $N$} {\small \Comment{Batch size}}
			\State $(u_n,v_n)  \gets f_\phi(x_n)$
			\State $\tilde{x}_n \gets$ $f_\phi^{-1}(u_n, \vec{0}_{D-d})$
			\EndFor
\State
compute $\ell$
\State $\phi \gets \phi - \alpha\nabla_\phi\ell$
			\EndWhile
		
		\EndProcedure
		\\
		\Procedure{\underline{Generation}}{}
		\\
		{\textbf{output:} $\tilde{x} \in \mathbb{R}^D$}
		\State $u  \sim p_u(u) = N(0,I_d)$
		\State $\tilde{z} \gets (u, \vec{0}_{D-d})$
        \State $\tilde{x} \gets f^{-1}(\tilde{z})$
		\EndProcedure
	\end{algorithmic}
	\end{multicols}
\end{algorithm}

\subsection{Hierarchical learning}
\label{sec:hierarchical}
NFs assume that their transformed space is Gaussianized, But this assumption is not true in high dimensions. 
Since fitting a Gaussian distribution to the entire high-dimensional transformed data space is difficult, we propose a hierarchical learning procedure which tries to concentrate on the sub-manifolds of data. 
For this purpose, we pass the data through several single-step sub-manifold(s) leaning NFs like the pixel rejection method.
We fix the previously learned manifold in each step and apply the subsequent single-step manifold learning to it.
In other words, we first find a sub-manifold of data and then project data on this manifold.
This procedure is iteratively repeated.
This iterative algorithm can be seen as similar to Gaussianization \cite{chen2000gaussianization}, where they suggest an iterative algorithm to solve the density estimation problem in high dimensions. 
Therefore, we will gradually improve the likelihood (density estimation) as long as the reconstruction loss (or manifold learning) is acceptable.
All in all, an advantage of the proposed method is that it can have a sequence of such blocks that are trained iteratively. In other words, we also consider the hierarchy in training in addition to the hierarchy in the structure.
The pseudo-code of the hierarchical structure method is provided in \AlgoRef{algo:multi}.
\begin{algorithm}[h]
	\caption{The pseudo-code of the hierarchical dimension reduction method.}
	\label{algo:multi}

\begin{algorithmic}[1]
		\Procedure{\underline{Training}}{}
		\\
            {\textbf{Require:}
			Number of manifolds $M$, Manifold dimensions $d_1,..., d_k$, Huber functions thresholds $\delta_1,...,\delta_k$, Penalization terms coefficients $\lambda_1,...,\lambda_k$, Learning rates $\alpha_1,...,\alpha_k$,
			data $X=\{x_n\}_{n=1}^{n=N} \sim p_X(x)$.
            }
            \State $X_0 = X$
            \For{$m \gets 1$ to $M$}
            \State $\phi_{m}^* = $ PixelRejetion$(d_m, \delta_m, \lambda_m, \alpha_m, X_{m-1})$
            \Comment{\small train a single-step dimension reduction }
            \State $X_m \gets f_{\phi_{m}}^{-1}(X_{m-1})$ \Comment{\small project $X_{m-1}$ to $\mathcal{M}_m$}
			\EndFor
		\EndProcedure
		\\
		\Procedure{\underline{Generation}}{}
		\\
		{\textbf{output:} $\tilde{x} \in \mathbb{R}^D$}
		\State $u  \sim p_{u_M}(u) = N(0,I_{d_M})$
		\For{$m \gets M-1$ to $2$}
		    \State $u \gets f_{\phi_m}^{-1}(u, \vec{0}_{d_{m-1}-d{m}})$
		\EndFor
		\State $\tilde{x} \gets f_{\phi_1}^{-1}(u)$
		\EndProcedure
	\end{algorithmic}
\end{algorithm}
\section{Results}
\label{sec:results}
In this section, the results of each of the two proposed methods are presented in \SectionRef{sec:single-step-experiments} and \SectionRef{sec:hierarchical-experiments}.
Due to page limitations, experiments on another dataset, information about architecture, and hyper-parameters are available in the appendix.
\subsection{Single-step method}
\label{sec:single-step-experiments}
\begin{figure}
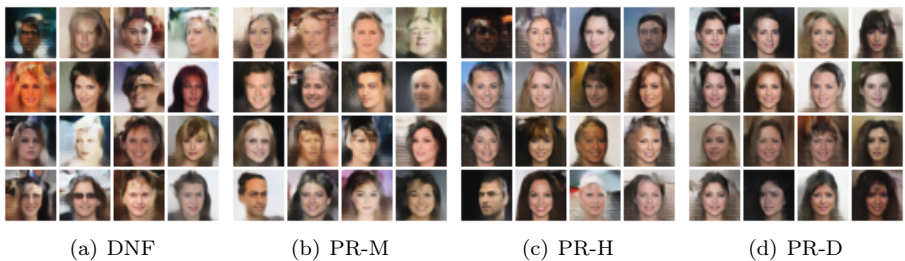

\centering
\subfigure[DNF]{\label{fig:vdirect_O}\includegraphics[width=0.24\textwidth]{\FigPath{comparison/DNF.png}}}
\subfigure[PR-M]{\label{fig:vdirect_R}\includegraphics[width=0.24\textwidth]{\FigPath{comparison/PR-M.png}}}
\subfigure[PR-H]{\label{fig:vdirect_R}\includegraphics[width=0.24\textwidth]{\FigPath{comparison/PR-H.png}}}
\subfigure[PR-D]{\label{fig:vdirect_R}\includegraphics[width=0.24\textwidth]{\FigPath{comparison/PR.png}}}
\caption{
Generated CelebA images for single-step methods corresponding to experiments in \TableRef{tab:scores1}.
}
\label{fig:scores1}
\end{figure}

The single-step method results for CelebA dataset are shown in \TableRef{tab:scores1} and \FigRef{fig:scores1}, respectively.
As discussed earlier, among related research in the literature, only DNF has a single-step training manifold learning and density estimation for NFs with a tractable likelihood.
We choose all its backbone models the same as ours in all experiments for a fair comparison.
Briefly, we employ two blocks together (a three layers (each having 32 flows) Glow model \cite{kingma2018glow} followed by a three layers RealNVP model \cite{dinh2016density}) as an NF to be more comparable in structure with DNF.
Therefore, the two-block single-step method means using a dimension preserving NF (RealNVP in our setting) to estimate the density of the manifold part instead of Gaussian distribution.
Also, all input images are resized to 32$\times$32 images.
Based on the reported results for CelebA dataset by DNF, we also consider $|\mathcal{M}|=500$. 
More details of backbone models are presented in the appendix.

The main goal of the single-step method is to find the main sub-manifold for generating data as long as it does not decrease the likelihood of data.
The reported results in \TableRef{tab:scores1} confirm that our single-step methods (PR-H: two-blocks pixel rejection method penalized by a Huber function,
PR-M: two-blocks pixel rejection method penalized by MSE, and
PR-D: one block (only Glow without RealNVP) pixel rejection method penalized by a Huber function)
outperform DNF in terms of BPD, whereas their generated images look visually similar, according to \FigRef{fig:scores1}.
Moreover, since the Huber function leads off-manifold pixels are penalized with a linear function instead of a quadratic one, it was to be expected that the MSE be greater in the penalized models with the Huber function.
%
Another strength is that the proposed method PR-D achieves better (or same as) images to DNF and other two-blocks proposed method, with fewer parameters while maintaining the likelihood.
As mentioned above, PR-D has only one block (Glow not followed by RealNVP).

\begin{table}
	\caption{
The best BPD/MSE scores of single-step methods
for data lie in $\mathcal{M} \subset \mathbb{R}^{500}$.
Dimension changing of the one-block (PR-D) and two-block (DNF, PR-M, PR-H) methods are $\mathbb{R}^{3072} \rightarrow \mathbb{R}^{500}$ and $\mathbb{R}^{3072} \rightarrow \mathbb{R}^{500} \rightarrow \mathbb{R}^{500}$.
\#Params means the number of trainable parameters.
}
	\label{tab:scores1}
	\centering
		
		

%
		
		

\begin{tabular}{lcccc}
		\toprule
		
		criterion
		& DNF
		& PR-M
		& PR-H
		& PR-D
		\\
		
		\cmidrule{1-5}\morecmidrules\cmidrule{1-5}
		MSE & 0.004 $\pm$ 0.0002  &   0.005 $\pm$  0.0003 &  0.01 $\pm$ 0.001 & 0.02 $\pm$ 0.001 \\
		BPD & 3.98 $\pm$ 0.03 & 3.52 $\pm$ 0.04 & 3.51 $\pm$ 0.04 & 3.52 $\pm$ 0.04 \\
		\#Params & $\approx$ 59M & $\approx$ 59M & $\approx$ 59M & $\approx$ 44M \\

		\bottomrule
\end{tabular}
\end{table}

\begin{figure}
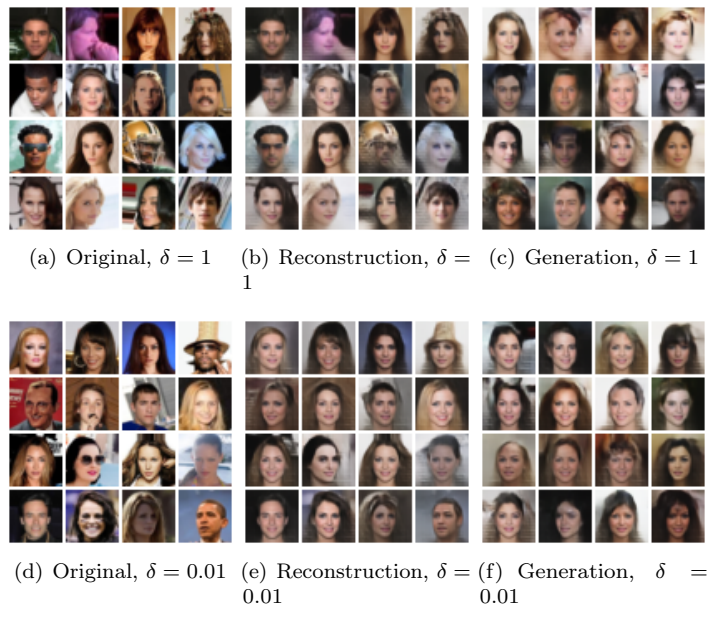

\centering
\subfigure[Original, $\delta=1$]{\label{fig:v26_O}\includegraphics[width=0.25\textwidth]{\FigPath{Delta/v26_O_19.png}}}
\subfigure[Reconstruction, $\delta=1$]{\label{fig:v26_R}\includegraphics[width=0.25\textwidth]{\FigPath{Delta/v26_R_19.png}}}
\subfigure[Generation, $\delta=1$]{\label{fig:v26_G}\includegraphics[width=0.25\textwidth]{\FigPath{Delta/v26_G_51.png}}}
\\
\subfigure[Original, $\delta=0.01$]{\label{fig:v28_O}\includegraphics[width=0.25\textwidth]{\FigPath{Delta/v28_O_0.png}}}
\subfigure[Reconstruction, $\delta=0.01$]{\label{fig:v28_R}\includegraphics[width=0.25\textwidth]{\FigPath{Delta/v28_R_0.png}}}
\subfigure[Generation, $\delta=0.01$]{\label{fig:v28_G}\includegraphics[width=0.25\textwidth]{\FigPath{Delta/v28_G_0.png}}}
\caption{Original, reconstructed, and generated images by one-block single-step pixel rejection method (PR-D) with $|\mathcal{M}|=500$ for different Huber function parameter $\delta$.
}
\label{fig:threshold}
\end{figure}

The pixel rejection method can be considered as a pixel-level outlier detection method.
An interesting case in the pixel rejection method is the effect of
$\delta$
on results. 
The important point is that $\delta$ is a boundary transition from a quadratic penalization to a linear one.
The results of this experiment on the CelebA dataset are shown in \FigRef{fig:threshold}.
As observed from the results, reducing $\delta$ leads to a decrease in the accuracy of the reconstruction images, especially in noisy pixels (like pixels belong to the glasses), by not learning off-manifold pixels.
At the same time, the sample quality remains good.
%
\subsection{Hierarchical training method}
\label{sec:hierarchical-experiments}

In this section, we provide experiments on hierarchy in structure and training.
Hierarchy in training means sequentially training blocks of a multi-block model each step by freezing previously trained blocks, while hierarchy in the structure means having a multi-block structure with end-to-end training.
It is worth noting that all reported models in this section include two blocks (Glow followed by a RealNVP model).
Accordingly, the used abbreviations are introduced.
H-XY is a hierarchical training method in which the first and second models are penalized by X and Y loss functions, respectively.
As loss function suffixes (X or Y), M and H are used for MSE and Huber function.
EE-H and EE-M stand for end-to-end training with Huber function and MSE, respectively.
In other words, EE-H and EE-M have hierarchical structures without hierarchical training.

\begin{figure}[t]
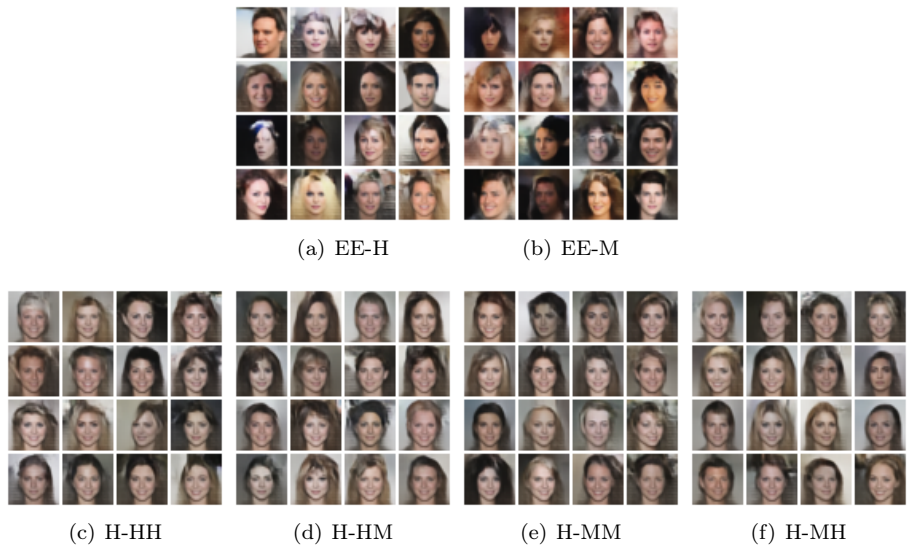

\centering
\subfigure[EE-H]{\label{fig:vTwoStep_O}\includegraphics[width=0.24\textwidth]{\FigPath{comparison/EE_H.png}}}
\subfigure[EE-M]{\label{fig:vdirect_G}\includegraphics[width=0.24\textwidth]{\FigPath{comparison/EE_M.png}}}
\\
\subfigure[H-HH]{\label{fig:vdirect_O}\includegraphics[width=0.24\textwidth]{\FigPath{comparison/H_HH.png}}}
\subfigure[H-HM]{\label{fig:vdirect_R}\includegraphics[width=0.24\textwidth]{\FigPath{comparison/H_HM.png}}}
\subfigure[H-MM]{\label{fig:vdirect_G}\includegraphics[width=0.24\textwidth]{\FigPath{comparison/H_MM.png}}}
\subfigure[H-MH]{\label{fig:vTwoStep_O}\includegraphics[width=0.24\textwidth]{\FigPath{comparison/H_MH.png}}}
\caption{
Generated CelebA images for hierarchical methods corresponding to reported experiments in \TableRef{tab:scores2} include hierarchical structure and hierarchical training, when $\mathbb{R}^{3072} \rightarrow \mathbb{R}^{1000} \rightarrow \mathbb{R}^{500}$.
}
\label{fig:scores2}
\end{figure}

%
Hierarchical method experiments for CelebA dataset with in a two-blocks hierarchical method ($\mathbb{R}^{3072} \rightarrow \mathbb{R}^{1000} \rightarrow \mathbb{R}^{500}$) are presented in \TableRef{tab:scores2} and \FigRef{fig:scores2}.
Based on the results, the differences in the hierarchical methods are negligible.
Still, the important point is that the likelihood improves considerably compared to single-step methods reported in \TableRef{tab:scores1}.
Still, the important point is that the sub-manifold likelihood improves considerably compared to single-step methods. 
Based on the reported results in \FigRef{fig:scores2}, it is exciting that in the hierarchical training method, the quality of the generated samples is much better than the end-to-end training (hierarchy in structure only) with the same architecture and the single-step methods (\FigRef{fig:scores1}).
All in all, it seems that the hierarchical training, not just a hierarchical structure, focuses on the main sub-manifold by removing off-manifold parts like the background.
Compared to the single-step results in \TableRef{tab:scores1}, the key point is that the hierarchical training methods prefer the generation quality to the reconstruction one.

\begin{table}[h]
	\caption{
    The best MSE/BPD scores of hierarchical methods, hierarchical structure (EE-H and EE-M), and hierarchical training (H-HH, H-HM, H-MM, and H-MH methods).
    These experiments are for a two-block model on CelebA dataset
    when
    $\mathbb{R}^{3072} \rightarrow \mathbb{R}^{1000} \rightarrow \mathbb{R}^{500}$.
    }
    \setlength{\tabcolsep}{2.5pt} 
	\label{tab:scores2}
	\centering
	\resizebox{\textwidth}{!}{

		
		
\begin{tabular}[\linewidth]{lcccccc}
		\toprule

		criterion
		& EE-H
		& EE-M
		& H-HH
		& H-HM
		& H-MM
		& H-MH\\
		
		\cmidrule{1-7}\morecmidrules\cmidrule{1-7}
		MSE & 0.02 $\pm$ 0.001 & 0.005 $\pm$ 0.0002 & 0.05 $\pm$ 0.004 & 0.05 $\pm$ 0.004 & 0.05 $\pm$ 0.004 & 0.05 $\pm$ 0.004 \\
		BPD & 3.53 $\pm$ 0.05 & 3.51 $\pm$ 0.05 &  3.51 $\pm$ 0.05   & 3.51 $\pm$ 0.05 & 3.47 $\pm$ 0.05 & 3.48 $\pm$ 0.05 \\
		
		\bottomrule
\end{tabular}
	}
\end{table}

\begin{figure}
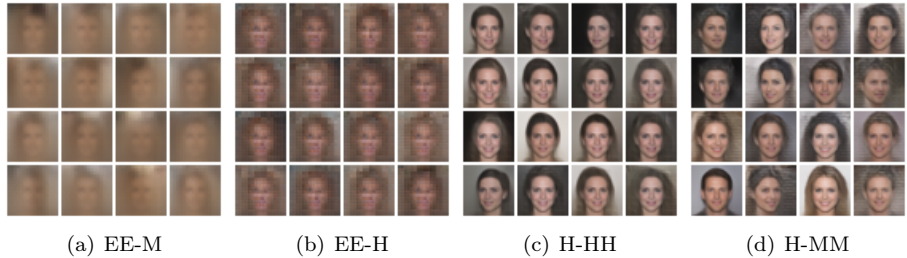

\centering
\subfigure[EE-M]{\label{fig:vdirect_G}\includegraphics[width=0.24\textwidth]{\FigPath{train_structure/EE_M_100.png}}}
\subfigure[EE-H]{\label{fig:vdirect_G}\includegraphics[width=0.24\textwidth]{\FigPath{train_structure/EE_H_100.png}}}
\subfigure[H-HH]{\label{fig:vdirect_G}\includegraphics[width=0.24\textwidth]{\FigPath{train_structure/H_HH_100.png}}}
\subfigure[H-MM]{\label{fig:vdirect_G}\includegraphics[width=0.24\textwidth]{\FigPath{train_structure/H_MM_100.png}}}
\caption{
Generated images by the hierarchical method (end-to-end: training means hierarchy in structure, hierarchical training: hierarchy in both structure and training) for significant dimension reduction of CelebA dataset, $\mathbb{R}^{3072} \rightarrow \mathbb{R}^{500} \rightarrow \mathbb{R}^{100}$.
}
\label{fig:TwoStep}
\end{figure}

Based on the above discussion, it is reinforced that hierarchical training methods for dimension reduction lead to better manifold learning compared to hierarchy in the structure (end-to-end training).
In order to further evaluate, the performance of the hierarchy in structure and the hierarchy in training for significantly dimension reduction (a two-blocks model, $\mathbb{R}^{3072} \rightarrow \mathbb{R}^{500} \rightarrow \mathbb{R}^{100}$) is presented in \FigRef{fig:TwoStep}.
The generated images indicate that the hierarchical training method helps prevent blurring.
In addition to the excellent performance of the hierarchical training method, optimizing the end-to-end training (only hierarchy in structure) for very low-dimensional manifolds such as $\mathcal{M} \subset \mathbb{R}^{100}$ is associated with instability. In contrast, hierarchical training helps to facilitate optimization.
On the contrary, the hierarchical training method's diversity of the generated images is reduced when the target manifold is embedded in a very low dimension.
In our opinion, this is a good achievement that the model learns a limited but perfectly sub-manifold.

\section{Conclusion}
In this paper, we pursued a solution to the problem of simultaneous manifold learning and density estimation using normalizing flows.
We proposed two new methods to address the mentioned problem, a single-step method (called pixel rejection) and a hierarchical method.
We used a per-pixel penalization function to penalize off-manifold pixels in the pixel rejection method.
The Huber function was employed here due to its transition mode from linear to quadratic. It causes to penalize off-manifold parts linearly, and the model focuses on learning the sub-manifold well.
Our experiments showed the capability of pixel rejection in finding a sub-manifold of data and estimating its likelihood.
pixel rejection can also generate samples similar to the latest state-of-the-art manifold learning in normalizing flows in terms of visual quality.
Furthermore, we introduced a new framework called hierarchical structure.
We considered two hierarchical approaches: hierarchy in structure which means that we can cascade different blocks of pixel rejection with each other and hierarchy in training which means that we can train different blocks in a hierarchical procedure.
Our results showed the superiority of these approaches compared to the single-step strategies
in terms of generating high-quality samples.
The latter is especially useful when we want to project high-dimensional data to a very lower-dimensional manifold.
We observed hierarchical training facilitates optimization in this case.

A primary assumption of the current study is embedding data on a manifold with one chart.
This assumption has also been used in similar works such as M-flow, DNF, and Rectangular flow.
However, it is not true, especially for high-dimensional data such as images. We want to extend a multi-manifold version of the pixel rejection method in future work.
Finding sub-manifolds instead of one is easy to develop in our framework, and we postpone it as future work.


\label{sec:conclusion}

\bibliography{manuscript}

\begin{thebibliography}{10}

\bibitem{brehmer2020flows}
J.~Brehmer and K.~Cranmer.
\newblock Flows for simultaneous manifold learning and density estimation.
\newblock In {\em Advances in Neural Information Processing Systems},
  volume~33, 2020.

\bibitem{caterini2021rectangular}
A.~L. Caterini, G.~Loaiza-Ganem, G.~Pleiss, and J.~P. Cunningham.
\newblock Rectangular flows for manifold learning.
\newblock In {\em Advances in Neural Information Processing Systems},
  volume~34, 2021.

\bibitem{chen2000gaussianization}
S.~Chen and R.~Gopinath.
\newblock Gaussianization.
\newblock In {\em Advances in Neural Information Processing Systems},
  volume~13, 2000.

\bibitem{cunningham2020normalizing}
E.~Cunningham, R.~Zabounidis, A.~Agrawal, I.~Fiterau, and D.~Sheldon.
\newblock Normalizing flows across dimensions.
\newblock In {\em International Conference on Machine Learning Workshop on
  Invertible Neural Networks, Normalizing Flows, and Explicit Likelihood
  Models}, 2021.

\bibitem{dinh2016density}
L.~Dinh, J.~Sohl-Dickstein, and S.~Bengio.
\newblock Density estimation using real {NVP}.
\newblock In {\em International Conference on Learning Representations}, 2017.

\bibitem{gemici2016normalizing}
M.~C. Gemici, D.~Rezende, and S.~Mohamed.
\newblock Normalizing flows on {Riemannian} manifolds.
\newblock {\em arXiv preprint arXiv:1611.02304}, 2016.

\bibitem{goodfellow2014generative}
I.~Goodfellow, J.~Pouget-Abadie, M.~Mirza, B.~Xu, D.~Warde-Farley, S.~Ozair,
  A.~Courville, and Y.~Bengio.
\newblock Generative adversarial nets.
\newblock In {\em Advances in neural information processing systems},
  volume~27, 2014.

\bibitem{horvat2021denoising}
C.~Horvat and J.-P. Pfister.
\newblock Denoising normalizing flow.
\newblock In {\em Advances in Neural Information Processing Systems},
  volume~34, 2021.

\bibitem{kalatzis2021multi}
D.~Kalatzis, J.~Z. Ye, J.~Wohlert, and S.~Hauberg.
\newblock Multi-chart flows.
\newblock {\em arXiv preprint arXiv:2106.03500}, 2021.

\bibitem{kim2020softflow}
H.~Kim, H.~Lee, W.~H. Kang, J.~Y. Lee, and N.~S. Kim.
\newblock Softflow: Probabilistic framework for normalizing flow on manifolds.
\newblock In {\em Advances in Neural Information Processing Systems},
  volume~33, 2020.

\bibitem{kingma2018glow}
D.~P. Kingma and P.~Dhariwal.
\newblock Glow: Generative flow with invertible 1x1 convolutions.
\newblock In {\em Advances in Neural Information Processing Systems},
  volume~31, 2018.

\bibitem{kingma2013auto}
D.~P. Kingma and M.~Welling.
\newblock Auto-encoding variational {Bayes}.
\newblock In {\em International Conference on Learning Representations}, 2014.

\bibitem{kothari2021trumpets}
K.~Kothari, A.~Khorashadizadeh, M.~de~Hoop, and I.~Dokmani{\'c}.
\newblock Trumpets: Injective flows for inference and inverse problems.
\newblock In {\em Uncertainty in Artificial Intelligence}, 2021.

\bibitem{liu2015faceattributes}
Z.~Liu, P.~Luo, X.~Wang, and X.~Tang.
\newblock Deep learning face attributes in the wild.
\newblock In {\em Proceedings of International Conference on Computer Vision
  (ICCV)}, December 2015.

\bibitem{mathieu2020riemannian}
E.~Mathieu and M.~Nickel.
\newblock Riemannian continuous normalizing flows.
\newblock In {\em Advances in Neural Information Processing Systems},
  volume~33, 2020.

\bibitem{murphy2022probabilistic}
K.~P. Murphy.
\newblock {\em Probabilistic machine learning: an introduction}.
\newblock MIT press, 2022.

\bibitem{rezende2015variational}
D.~Rezende and S.~Mohamed.
\newblock Variational inference with normalizing flows.
\newblock In {\em International Conference on Machine Learning}, pages
  1530--1538, 2015.

\bibitem{rezende2020normalizing}
D.~J. Rezende, G.~Papamakarios, S.~Racaniere, M.~Albergo, G.~Kanwar,
  P.~Shanahan, and K.~Cranmer.
\newblock Normalizing flows on tori and spheres.
\newblock In {\em International Conference on Machine Learning}, pages
  8083--8092, 2020.

\bibitem{ross2021conformal}
B.~L. Ross and J.~C. Cresswell.
\newblock Conformal embedding flows: Tractable density estimation on learned
  manifolds.
\newblock In {\em International Conference on Machine Learning Workshop on
  Invertible Neural Networks, Normalizing Flows, and Explicit Likelihood
  Models}, 2021.

\bibitem{pmlr-v37-sohl-dickstein15}
J.~Sohl-Dickstein, E.~Weiss, N.~Maheswaranathan, and S.~Ganguli.
\newblock Deep unsupervised learning using nonequilibrium thermodynamics.
\newblock In {\em International Conference on Machine Learning}, pages
  2256--2265, 2015.

\bibitem{theis2015note}
L.~Theis, A.~v.~d. Oord, and M.~Bethge.
\newblock A note on the evaluation of generative models.
\newblock In {\em International Conference on Learning Representations}, 2016.

\bibitem{yu2015lsun}
F.~Yu, A.~Seff, Y.~Zhang, S.~Song, T.~Funkhouser, and J.~Xiao.
\newblock Lsun: Construction of a large-scale image dataset using deep learning
  with humans in the loop.
\newblock {\em arXiv preprint arXiv:1506.03365}, 2015.

\end{thebibliography}
\bibliographystyle{abbrv}

\appendix
\label{sec:appendix}

Due to page limitations in the paper, 
the introduction of used datasets (\SectionRef{sec:datasets}), experiment setups (\SectionRef{sec:experiments_setups}), the rest of the experiments, and the experiments on another dataset (\SectionRef{sec:additional_experiments}) are provided as follows.

\section{Datasets}
\label{sec:datasets}

The two used datasets in the current study are introduced below.
It should be noted that we can increase the number of layers and parameters in proportion to the original image's dimensions. Still, due to resource allocation constraints, we had to change the size of the images to $32 \times 32$ in our experiments.

\subsubsection{CelebA dataset}
CelebFaces Attributes (CelebA\footnote{https://mmlab.ie.cuhk.edu.hk/projects/CelebA.html}) dataset \cite{liu2015faceattributes} contains 202,599 face images of 10,177 persons.
It covers backgrounds with different variations in addition to diversity in faces.
Our paper used 200782 randomly selected images from this dataset as a train set.

\subsubsection{LSUN dataset}
This collection\footnote{https://www.yf.io/p/lsun} includes a variety of large-scale images categorized into ten different classes \cite{yu2015lsun}.
In this study, we only used bedroom class images.
This class comprises more than 3 million images in the original dataset, but we used only 303125 of them\footnote{https://www.kaggle.com/datasets/jhoward/lsun\_bedroom}.
We considered 284128 of them as a train set.

\section{Experiments Setups}
\label{sec:experiments_setups}
The proposed methods architecture along with training settings are introduced in this section.
The implementation codes of the proposed methods are attached to this file.

\subsection{Single-step method}
\label{sec:single_step_appendix}
\textbf{Architectures: }
We considered four structures in this part of the experiment, DNF, PR-M, PR-H, and PR-D.
In the first three, two NFs (named $f_\phi$ and $f_\theta$) are used, and in the last structure, a single NF (named $f_\phi$) is used.
We choose the Glow model \cite{kingma2018glow} for $f_\phi$ and the RealNVP model \cite{dinh2016density} for $f_\theta$. 
Our Glow model has three blocks, each with 32 steps of the flow.
Each step consists of an actnorm layer, an invertible $1 \times 1$ convolution layer, and an affine coupling layer (see \cite{kingma2018glow} for the details of each layer). There is also a squeeze layer before each step and a split layer after each step.
The RealNVP model comprises six affine coupling layers and six masking layers before each affine coupling.


\textbf{Training:} 
All implementations are done in the PyTorch\footnote{https://pytorch.org/}(version 1.10.0+cu102) framework. We used Adam as our optimization algorithm with a learning rate of 1e-5 and a batch size of 64, and we set the value of the penalization term coefficient ($\lambda$) to one.
All training experiments were done on a single GPU (GTX 1080 Ti), with Cuda version 11.4 and 11019GB memory.


\subsection{Hierarchical method}
\textbf{Architectures: } For this method, we presented several experiments in the paper. All of them are composed of two NFs $f_\phi$ and $f_\theta$, whose architectures have already been introduced in the above section (\SectionRef{sec:single_step_appendix}).


\textbf{Training:} The best configuration for training the hierarchical method (hierarchy in structure or hierarchy in training) is similar to the single-step method.

\section{Additional experiments}
\label{sec:additional_experiments}
Due to page limitations, only experiments related to the CelebA dataset were reported in the paper.
This section presents the remaining experiments and new experiments on another dataset (LSUN).

Before providing the experiments, it should be noted that the FID score was not reported for CelebA dataset in the main paper, but we will report it in the following experiments for the LSUN dataset.
Despite many indicators measuring the quality of the images in generative models, no ideal one is not ideal like human perception.
In the case of CelebA dataset, although the quality of images was good visually, the evaluated FID (a Wasserstein-based metric for generative model) for the proposed methods was a little high. So, we did not report numerical results.

\subsection{The effect of the pixel rejection method on pixels}
\begin{figure}[H]
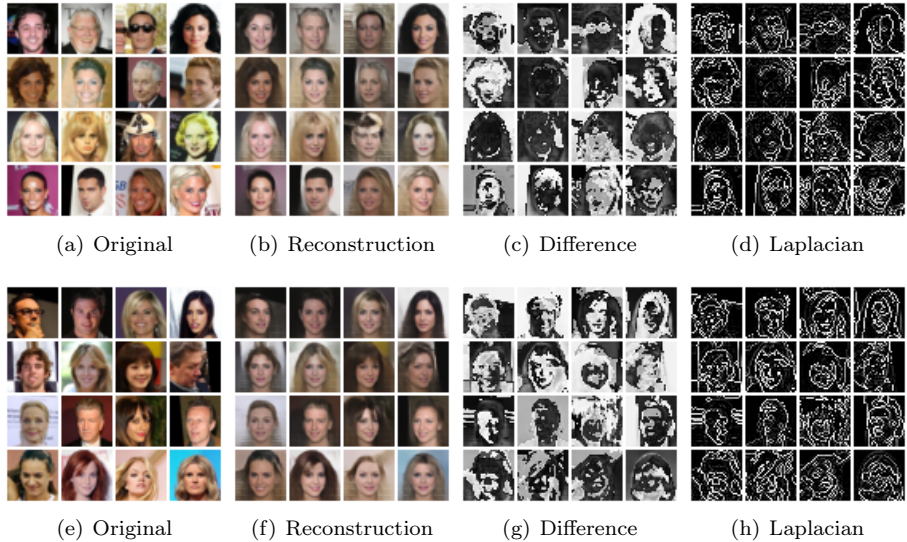

\centering
\subfigure[Original]{\label{fig:vdirect_O}\includegraphics[width=0.24\textwidth]{\FigPath{Difference/O_7.png}}}
\subfigure[Reconstruction]{\label{fig:vdirect_R}\includegraphics[width=0.24\textwidth]{\FigPath{Difference/R_7.png}}}
\subfigure[Difference]{\label{fig:vdirect_G}\includegraphics[width=0.24\textwidth]{\FigPath{Difference/D_7.png}}}
\subfigure[Laplacian]{\label{fig:vTwoStep_G}\includegraphics[width=0.24\textwidth]{\FigPath{Difference/E_7.png}}}
\subfigure[Original]{\label{fig:vdirect_O}\includegraphics[width=0.24\textwidth]{\FigPath{Difference/O_9.png}}}
\subfigure[Reconstruction]{\label{fig:vdirect_R}\includegraphics[width=0.24\textwidth]{\FigPath{Difference/R_9.png}}}
\subfigure[Difference]{\label{fig:vdirect_G}\includegraphics[width=0.24\textwidth]{\FigPath{Difference/D_9.png}}}
\subfigure[Laplacian]{\label{fig:vTwoStep_G}\includegraphics[width=0.24\textwidth]{\FigPath{Difference/E_9.png}}}
\caption{
Original, reconstruction, difference, and edges of difference CelebA images for reported PR-D with $d=500$ method in the paper.
The two rows contain randomly selected images from the results.
Each row includes the original image, the corresponding reconstruction, the difference between the original and reconstructed images, and the differences' edges.
}
\label{fig:Diff}
\end{figure}

In order to further intuition on the performance of the pixel rejection method, the differences between the original images and the reconstructed ones are illustrated in \FigRef{fig:Diff}.
Moreover, applying an edge detection method (Laplacian) to differences indicates the most important patterns.
As it is known, in images that have outlier pixels (like pixels that belong to glasses), outlier pixels removal is quite obvious in the difference between the two images.
Because our main goal was to improve the generation, we significantly differ between the original image and the reconstruction. Therefore, the details of the face in the original images are lost.
In other words, the proposed model seeks to learn the image's outline and does not learn the details (especially the details that are less present in the scene).

\subsection{LSUN (bedroom) dataset}
Previously reported results in the paper included a well-defined face manifold dataset. However, it is also necessary to evaluate the model's performance in crowded and cluttered manifolds. 
The class of bedrooms in the LSUN dataset is our choice for following experiments.

\subsubsection{Comparing hierarchy in structure and training}
The corresponding results for the LSUN dataset for different configurations such as hierarchy in structure (EE-M, EE-H), hierarchy in training (H-MM, H-HH), and DNF are presented in \TableRef{tab:loss_appendix1} and \FigRef{fig:fig2}.
All target manifolds lie in $\mathbb{R}^{1000}$.
The numerical result confirms that hierarchical structure leads to better manifold learning (less reconstruction error) in all cases, such as hierarchy in structure or hierarchy in training.
Moreover, its density estimation can reach a better likelihood in terms of BPD.
It seems that hierarchical training is prone to better density estimation compared to end-to-end training.
However, the FID score indicates a slight superiority of the DNF method.
This is not unexpected because our loss function penalizes diversity, as discussed in the main paper.
However, our FID scores are also relatively good.

\begin{table}
	\caption{The best FID, MSE, and BPD scores for hierarchical methods (hierarchy in structure and training) and DNF on the LSUN dataset for target manifold embedded in $\mathbb{R}^{1000}$.}
	\label{tab:loss_appendix1}
	\setlength{\tabcolsep}{\tabcolsep}
	\centering
		
		


\resizebox{\textwidth}{!}
{
\begin{tabular}{l|cc|cc|cc}
		\toprule
		
		\rotatebox{90}{\tiny criterion}
		& \makecell{EE-M \\ {\tiny ($3072 \rightarrow 2000 \rightarrow 1000$)}}
		& \makecell{EE-H \\ {\tiny ($3072 \rightarrow 2000 \rightarrow 1000$)}}
		& \makecell{H-MM \\ {\tiny ($3072 \rightarrow 2000 \rightarrow 1000$)}}
		& \makecell{H-HH \\ {\tiny ($3072 \rightarrow 2000 \rightarrow 1000$)}}
		& \makecell{DNF \\ {\tiny ($3072 \rightarrow 1000 \rightarrow 1000$)}}
		\\
		
		\cmidrule{1-6}\morecmidrules\cmidrule{1-6}
 		FID & 63.96 & 82.5 & 64.40 & 71.53 & 55.45 \\
		MSE & 0.0003 $\pm$ 9e-6 & 0.0008 $\pm$ 2e-5 & 0.0007 $\pm$ 2e-5 & 0.0008 $\pm$ 2e-5 & 0.001 $\pm$ 9e-5 \\
		BPD & 3.65 $\pm$ 0.05 & 3.7 $\pm$ 0.05 & 3.62 $\pm$ 0.0 & 3.62 $\pm$ 0.05 & 4.10 $\pm$ 0.04 \\

		\bottomrule
\end{tabular}
}
		
\end{table}

\begin{figure}
\centering
\subfigure[EE-M, $\mathcal{M} \subset \mathbb{R}^{1000}$]{\label{fig:vdirect_O}\includegraphics[width=0.45\textwidth]{\FigPath{appendix/EE_M_1000.png}}}
\subfigure[EE-H, $\mathcal{M} \subset \mathbb{R}^{1000}$]{\label{fig:vdirect_R}\includegraphics[width=0.45\textwidth]{\FigPath{appendix/EE_H_1000.png}}}
\\
\subfigure[H-MM, $\mathcal{M} \subset \mathbb{R}^{1000}$]{\label{fig:vdirect_O}\includegraphics[width=0.45\textwidth]{\FigPath{appendix/H_MM_1000.png}}}
\subfigure[H-HH, $\mathcal{M} \subset \mathbb{R}^{1000}$]{\label{fig:vdirect_R}\includegraphics[width=0.45\textwidth]{\FigPath{appendix/H_HH_1000.png}}}
\\
\subfigure[DNF, $\mathcal{M} \subset \mathbb{R}^{1000}$]{\label{fig:vdirect_R}\includegraphics[width=0.45\textwidth]{\FigPath{appendix/DNF_1000.png}}}
\caption{
The randomly generated images for corresponding reported experiments in \TableRef{tab:loss_appendix1}.
}
\label{fig:fig2}
\end{figure}

In the hierarchical approach, sub-manifolds in each stage should be well-learned. So, adjusting $\lambda$'s value is crucial. 
Instead of varying $\lambda$, we can multiply the likelihood term in our loss function by a coefficient. 
The smaller coefficient value means that the model focuses on manifold learning (reconstruction error).
We examine the effect of this coefficient value on the LSUN (bedroom) dataset for two values of 0.5 and 1 for density estimation and manifold learning, respectively.
It should be noted that we have used this ccoefficient only in the loss function of the first NF ($f_\phi$), and for the second NF ($f_\theta$), we have always considered this value equal to 1.
The corresponding results of this experiment are available in \TableRef{tab:loss_appendix2} and \FigRef{fig:loss_appendix2}.
Based on the results, the FID score is improved compared to corresponding previous results reported in \TableRef{tab:loss_appendix1}.
An important point about changing the proportion of likelihood and penalization in the first NF in hierarchical training is that graduation helps to adjust the manifold learning process.
In other words, to avoid the complexity of the joint cost function (equal proportion for likelihood and penalization), it is sufficient to estimate the relative density at each step while being precise with the manifold learning in that step.

\begin{table}[H]
	\caption{
	The best FID, MSE, and BPD scores for hierarchical training in case of changing the proportion of likelihood and penalization (0.5 to 1) on the LSUN dataset when the target manifold is embedded in $\mathbb{R}^{1000}$.
	$ ^\star$ means changed ratio in the first module.}
	\label{tab:loss_appendix2}
	\setlength{\tabcolsep}{\tabcolsep}
	\centering
%
%
%
%
%
%
%
%
\begin{tabular}{l|ccc}
		\toprule
		
		criterion
		& \makecell{H-MM$^\star$ \\ {\tiny ($3072 \rightarrow 2000 \rightarrow 1000$)}}
		& \makecell{H-MH$^\star$ \\ {\tiny ($3072 \rightarrow 2000 \rightarrow 1000$)}}
		\\
		
		\cmidrule{1-3}\morecmidrules\cmidrule{1-3}
 		FID & 62.72 & 62.79  \\
		MSE & 0.0007 $\pm$ 2e-5 & 0.0007 $\pm$ 2e-5  \\
		BPD & 3.62 $\pm$ 0.05 & 3.62 $\pm$ 0.05 \\

		\bottomrule
\end{tabular}
\end{table}

\begin{figure}[H]
\centering
\subfigure[H-MM$^*$, $\mathcal{M} \subset \mathbb{R}^{1000}$]{\label{fig:vdirect_O}\includegraphics[width=0.45\textwidth]{\FigPath{appendix/H_MM_1000_star.png}}}
\subfigure[H-MH$^*$, $\mathcal{M} \subset \mathbb{R}^{1000}$]{\label{fig:vdirect_R}\includegraphics[width=0.45\textwidth]{\FigPath{appendix/H_MH_1000_star.png}}}
\caption{
The randomly generated images for corresponding reported experiments in \TableRef{tab:loss_appendix2}.
}
\label{fig:loss_appendix2}
\end{figure}

\subsection{The effect of manifold's dimension on the LSUN dataset}
The target manifold in the previous experiment is in the 1000-dimensional space. Although the LSUN dataset is more complex than CelebA, it seems possible to reach a manifold with a smaller dimension.
The results of this experiment are presented in \TableRef{tab:loss_appendix4} and \FigRef{fig:fig5}.
The important point is that the proposed methods, regardless of their configuration, achieve better results than DNF and the previous experiments in terms of numerical and visual results.
Moreover, this experiment confirms that the proposed framework, despite its structural similarity to DNF, leads to relatively better results in the case of lower dimensions.

\begin{table}[H]
	\caption{The best FID, MSE, and BPD scores for hierarchical structure and DNF on the LSUN dataset for target manifold embedded in $\mathbb{R}^{700}$.}
	\label{tab:loss_appendix4}
	\setlength{\tabcolsep}{\tabcolsep}
	\centering
\begin{tabular}{l|cc|cc}
		\toprule
		
		criterion
		& \makecell{EE-M \\ {\tiny ($3072 \rightarrow 1000 \rightarrow 700$)}}
		& \makecell{EE-H \\ {\tiny ($3072 \rightarrow 1000 \rightarrow 700$)}}
		& \makecell{DNF \\ {\tiny ($3072 \rightarrow 700 \rightarrow 700$)}}
		\\
		
		\cmidrule{1-4}\morecmidrules\cmidrule{1-4}
 		FID & 62.01 & 71.56 & 80.43 \\
		MSE & 0.0004 $\pm$ 1e-5 & 0.0009 $\pm$ 2e-5 & 0.003 $\pm$ 0.0002 \\
		BPD & 3.66 $\pm$ 0.06 & 3.62 $\pm$ 0.06 & 4.14 $\pm$ 0.04 \\

		\bottomrule
\end{tabular}
\end{table}

\begin{figure}[H]
\centering
\subfigure[EE-M, $\mathcal{M} \subset \mathbb{R}^{700}$]{\label{fig:vdirect_O}\includegraphics[width=0.45\textwidth]{\FigPath{appendix/EE_M_700.png}}}
\subfigure[EE-H, $\mathcal{M} \subset \mathbb{R}^{700}$]{\label{fig:vdirect_R}\includegraphics[width=0.45\textwidth]{\FigPath{appendix/EE_H_700.png}}}
\\
\subfigure[DNF, $\mathcal{M} \subset \mathbb{R}^{700}$]{\label{fig:vdirect_O}\includegraphics[width=0.45\textwidth]{\FigPath{appendix/DNF_700.png}}}
\caption{
The randomly generated images for corresponding reported experiments in \TableRef{tab:loss_appendix4}.
}
\label{fig:fig5}
\end{figure}

\end{document}


\appendix
\section{Appendix}
\label{sec:appendix}

\subsection{Related works}



\subsection{Architecture and hyper-parameter}


\subsubsection{pixel/sample rejection}
\\
\textbf{Architecture: } We choose the Glow model as the casestudy for the proposed method for the image dataset.
Summarily, it can be considered as a convolutional extension of multi-scale RealNVP.
The model consists of several consecutive blocks that part of the output of each block goes directly to the next layer and part of it goes directly to the latent space.
In other words, the hidden space is formed gradually.
Unlike the state-of-the-art Glow structure for CelebA dataset, we choose a simple design with fewer blocks.
Since we resize images as 32 $ \times $ 32, we simplify the network to have 3 blocks, each with 32 flows.
\\
\textbf{Hyper-parameter:} The implementation was done in the PyTorch\footnote{https://pytorch.org/} framework. Adam optimized it with a learning rate of 1e-5 for a batch size of 64.

\subsubsection{Hierarchical dimension reduction}
\\
\textbf{Architecture: } In this method, a pre-trained model (exactly the same as the pixel/sample rejection) is initially used for transforming data.
After that, another flow is employed on the part of the transformed data corresponding to the manifold.
A RealNVP model is used as a case study for the second normalizing flow.
Briefly, we introduce RealNVP. It is a pioneer coupling-based normalizing flow that splits the feature vectors into two parts.
One part is directly transferred to the next level, and another is transferred by a parametric function conditioned on the first part.
Two neural networks usually make conditioning.
We use the same architecture for these functions as the three consecutive linear layers, each with 256 hidden units followed by the non-linear activation function ReLU. We consider the composition of 8 coupling layers as baseline normalizing flow.
\\
\textbf{Hyper-parameter:} The model was implemented in PyTorch and optimized by Adam and a learning rate of 1e-5 for batch size 64.
According to the proposed method, we need more hyper-parameter tuning like the coefficient of the regularization term.
We set it 1 in all reported experiments.

\subsection{Dataset}

\subsubsection{CelebA\footnote{https://mmlab.ie.cuhk.edu.hk/projects/CelebA.html} dataset}
CelebFaces Attributes (CelebA) dataset contains 202,599 face images of 10,177 persons.
It covers backgrounds with different variations in addition to diversity in faces.

\subsubsection{LSUN\footnote{https://www.yf.io/p/lsun} dataset}
This collection includes a variety of large-scale images categorized into 10 different classes.
In this study, we use only a part of one of its classes called the bedrooms.
This class contains 3,033,342 images, and we only use a quarter of them.

\subsection{Extra experiments}
\textbf{The effect of Pixel rejection on pixels:}
In order to further intuition on the performance of the pixel rejection method, the differences between the original images and the reconstructed ones are illustrated in \FigRef{fig:Diff}.
Moreover, applying an edge detection method (Laplacian) to differences indicates the most important patterns.
As it is known, in images that have outlier pixels (like the presence of glasses), outlier pixels removal is quite obvious in the difference between the two images.
Because our main goal was to improve the generation, we significantly differ between the original image and the reconstruction. Therefore, the details of the face in the original images are lost.
In other words, the proposed model seeks to learn the image's outline and does not learn the details (especially the details that are less present in the scene).
\begin{figure}[H]
\centering
\subfigure[Original]{\label{fig:vdirect_O}\includegraphics[width=0.2\textwidth]{\FigPath{Difference/O_7.png}}}
\subfigure[Reconstruction]{\label{fig:vdirect_R}\includegraphics[width=0.2\textwidth]{\FigPath{Difference/R_7.png}}}
\subfigure[Difference]{\label{fig:vdirect_G}\includegraphics[width=0.2\textwidth]{\FigPath{Difference/D_7.png}}}
\subfigure[Laplacian]{\label{fig:vTwoStep_G}\includegraphics[width=0.2\textwidth]{\FigPath{Difference/E_7.png}}}
\subfigure[Original]{\label{fig:vdirect_O}\includegraphics[width=0.2\textwidth]{\FigPath{Difference/O_9.png}}}
\subfigure[Reconstruction]{\label{fig:vdirect_R}\includegraphics[width=0.2\textwidth]{\FigPath{Difference/R_9.png}}}
\subfigure[Difference]{\label{fig:vdirect_G}\includegraphics[width=0.2\textwidth]{\FigPath{Difference/D_9.png}}}
\subfigure[Laplacian]{\label{fig:vTwoStep_G}\includegraphics[width=0.2\textwidth]{\FigPath{Difference/E_9.png}}}
\caption{Original, reconstruction, difference, and edges of difference image. 
The two rows contain randomly selected images from the results.
Each row includes the original image, the corresponding reconstruction, the difference between the original and reconstructed images, and the differences' edges.
The embedded manifold in this experiment for CelebA is $\mathbb{R}^{500}$.
}
\label{fig:Diff}
\end{figure}
\\
\\
\noindent
\textbf{The performance on natural and crowded images:}
Previously reported results included a well-defined face manifold dataset. However, it is also necessary to evaluate the model's performance in crowded and cluttered manifolds. 
The class of bedrooms in the LSUN dataset is our choice for this experiment.
The reported results in \FigRef{fig:lsun} confirm the excellent performance of the method.
Outlier pixels in this dataset are not easily definable. But what is clear from the results is that some details and patterns of objects are considered outliers while the original format of the object remains preserved.
\begin{figure}[H]
\centering
\subfigure[Original]{\label{fig:vdirect_O}\includegraphics[width=0.3\textwidth]{\FigPath{ReconstructionLSUN/O_0.png}}}
\subfigure[Reconstruction]{\label{fig:vdirect_R}\includegraphics[width=0.3\textwidth]{\FigPath{ReconstructionLSUN/R_0.png}}}
\subfigure[Generation]{\label{fig:vdirect_G}\includegraphics[width=0.3\textwidth]{\FigPath{ReconstructionLSUN/G_0.png}}}
\subfigure[Original]{\label{fig:vdirect_O}\includegraphics[width=0.3\textwidth]{\FigPath{ReconstructionLSUN/O_1.png}}}
\subfigure[Reconstruction]{\label{fig:vdirect_R}\includegraphics[width=0.3\textwidth]{\FigPath{ReconstructionLSUN/R_1.png}}}
\subfigure[Generation]{\label{fig:vdirect_G}\includegraphics[width=0.3\textwidth]{\FigPath{ReconstructionLSUN/G_1.png}}}
\subfigure[Original]{\label{fig:vdirect_O}\includegraphics[width=0.3\textwidth]{\FigPath{ReconstructionLSUN/O_6.png}}}
\subfigure[Reconstruction]{\label{fig:vdirect_R}\includegraphics[width=0.3\textwidth]{\FigPath{ReconstructionLSUN/R_6.png}}}
\subfigure[Generation]{\label{fig:vdirect_G}\includegraphics[width=0.3\textwidth]{\FigPath{ReconstructionLSUN/G_6.png}}}
\caption{
Original, reconstruction, and generated images of LSUN dataset (bedrooms class).
The learned embedded manifold in this experiment lies in $\mathbb{R}^{1000}$.
}
\label{fig:lsun}
\end{figure}
\\
\\
\noindent
\textbf{Exploring the effect of non-\textit{i.i.d} samples on sample rejection method:}
As mentioned in the proposed section, one of the advantages of the sample rejection method is the ability not to model off-manifold data distribution. In this regard, two different datasets (CelebA, LSUN) are used to understand these intuitions.
Therefore, the used dataset does not satisfy the \textit{identical} constraint in the \textit{i.i.d} definition.
The ratio of LSUN to CelebA is 1 to 3.
This combination causes some samples to be considered outliers, unlike the pixel rejection method, in which pixels can be outliers.
The following image is the performance of the trained model for two consecutive epochs.
As seen in the reconstruction images, the LSUN that has the least contribution to the training is not successful in the reconstruction.
Moreover, the generated samples mainly visualize the CelebA dataset that has more contribution to training.
\begin{figure}[H]
\centering
\subfigure[Original]{\label{fig:vdirect_O}\includegraphics[width=0.3\textwidth]{\FigPath{Merge/original8.png}}}
\subfigure[Reconstruction]{\label{fig:vdirect_R}\includegraphics[width=0.3\textwidth]{\FigPath{Merge/reconstruction8.png}}}
\subfigure[Generation]{\label{fig:vdirect_G}\includegraphics[width=0.3\textwidth]{\FigPath{Merge/sample8.png}}}
\subfigure[Original]{\label{fig:vdirect_O}\includegraphics[width=0.3\textwidth]{\FigPath{Merge/original9.png}}}
\subfigure[Reconstruction]{\label{fig:vdirect_R}\includegraphics[width=0.3\textwidth]{\FigPath{Merge/reconstruction9.png}}}
\subfigure[Generation]{\label{fig:vdirect_G}\includegraphics[width=0.3\textwidth]{\FigPath{Merge/sample9.png}}}
\caption{
The results of two consecutive epochs of training of non-\textit{i.i.d} dataset (combination of CelebA and LSUN),
original, reconstruction, and generated images. The embedded manifold lies in $\mathbb{R}^{500}$. 
}
\label{fig:Merge}
\end{figure}
\\
\\
\noindent
\textbf{Effect of $\delta$ on sample rejection: }
Another experiment is performed for the sample rejection method in this section.
In this experiment, the Huber threshold ($\delta$) is defined as the median of the batch loss function.
Therefore a group of samples with a loss more than the median is considered outliers.
They are scaled to get away from the threshold.
The results of this experiment are shown in the figure below.
It is clear that the outlier samples are linearly penalized with less impact with a small scale factor.
While on/nearly-on manifold samples are quadratically penalized.
The results show that the smaller scale leads to low diversity and high-quality generation. However, the large scale has more diversity.
\begin{figure}[H]
\centering
\subfigure[Original]{\label{fig:vdirect_O}\includegraphics[width=0.3\textwidth]{\FigPath{Extreme_low/original19.png}}}
\subfigure[Reconstruction]{\label{fig:vdirect_R}\includegraphics[width=0.3\textwidth]{\FigPath{Extreme_low/reconstruction19.png}}}
\subfigure[Generation]{\label{fig:vdirect_G}\includegraphics[width=0.3\textwidth]{\FigPath{Extreme_low/sample19.png}}}
\subfigure[Original]{\label{fig:vdirect_O}\includegraphics[width=0.3\textwidth]{\FigPath{Extreme_high/original19.png}}}
\subfigure[Reconstruction]{\label{fig:vdirect_R}\includegraphics[width=0.3\textwidth]{\FigPath{Extreme_high/reconstruction19.png}}}
\subfigure[Generation]{\label{fig:vdirect_G}\includegraphics[width=0.3\textwidth]{\FigPath{Extreme_high/sample19.png}}}
\caption{
Original, reconstruction, and generated images of CelebA in sample rejection method.
The first row is the results of small scale value, and the second row is the corresponding epoch result for large scale value.
The dimension of the manifold is $\mathbb{R}^{500}$.
}
\label{fig:LH}
\end{figure}